\documentclass[10pt, conference, compsocconf]{IEEEtran}

\usepackage{graphicx}
\usepackage{amsmath}
\usepackage{amssymb}
\usepackage{booktabs}
\usepackage{algorithm}
\usepackage{multirow}
\usepackage{algpseudocode}
\usepackage{makecell}
\usepackage{lipsum}
\usepackage{amsmath}
\usepackage{wrapfig}
\usepackage{bm}

\algnewcommand\algorithmicforeach{\textbf{for each}}
\algdef{S}[FOR]{ForEach}[1]{\algorithmicforeach\ #1\ \algorithmicdo}

\ifCLASSINFOpdf
\else
\fi
\begin{document}
%
\title{Meta Episodic learning with Dynamic Task Sampling for CLIP-based Point Cloud Classification}


\author{\IEEEauthorblockN{Shuvozit Ghose}
\IEEEauthorblockA{Department of Computer Science\\
University of Manitoba\\
Winnipeg, Canada\\
shuvozit.ghose@gmail.com}
\and
\IEEEauthorblockN{Yang Wang}
\IEEEauthorblockA{Department of Computer Science and Software Engineering\\
 Concordia University\\
Montreal, Canada\\
yang.wang@concordia.ca}
}


%


\maketitle

\begin{abstract}
Point cloud classification refers to the process of assigning semantic labels or categories to individual points within a point cloud data structure. Recent works have explored the extension of pre-trained CLIP to 3D recognition. In this direction, CLIP-based point cloud models like PointCLIP, CLIP2Point have become state-of-the-art methods in the few-shot setup. Although these methods show promising performance for some classes like airplanes, desks, guitars, etc, the performance for some classes like the cup, flower pot, sink, nightstand, etc is still far from satisfactory. This is due to the fact that the adapter of CLIP-based models is trained using randomly sampled N-way K-shot data in the standard supervised learning setup. In this paper, we propose a novel meta-episodic learning framework for CLIP-based point cloud classification, addressing the challenges of limited training examples and sampling unknown classes. Additionally, we introduce dynamic task sampling within the episode based on performance memory. This sampling strategy effectively addresses the challenge of sampling unknown classes, ensuring that the model learns from a diverse range of classes and promotes the exploration of underrepresented categories. By dynamically updating the performance memory, we adaptively prioritize the sampling of classes based on their performance, enhancing the model's ability to handle challenging and real-world scenarios. Experiments show an average performance gain of 3-6\% on ModelNet40 and ScanobjectNN datasets in a few-shot setup.

\end{abstract}


%
\IEEEpeerreviewmaketitle

\section{Introduction}
\begin{figure}[t]
\begin{center}
\includegraphics[width=.70\linewidth]{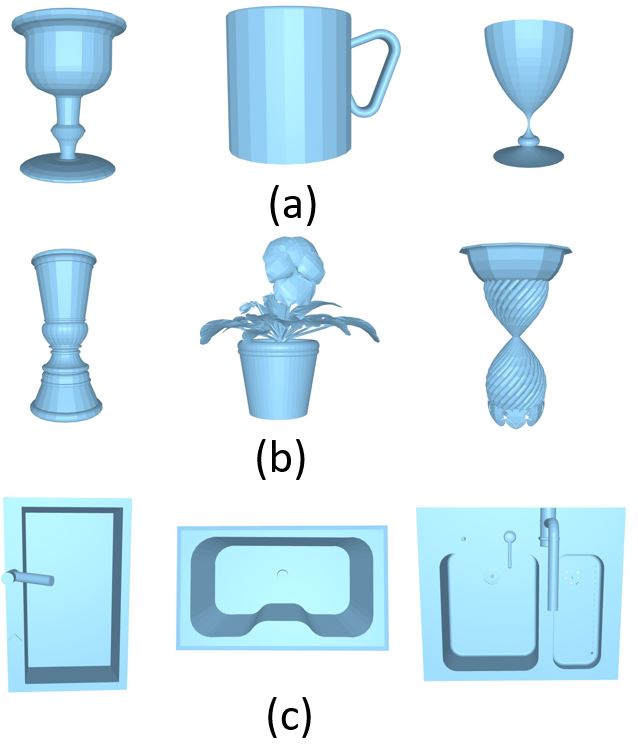}
\end{center}
    \vspace{-0.30cm}
  \caption{(a) cup (side view) (b) flower pot (side view) (c) sink (top view). We visualize the 3d object of the corresponding point cloud for better visual understanding.} 
  \vspace{-0.30cm}
\label{fig: Fig1}
\vspace{-0.30cm}
\end{figure}

Point cloud classification refers to the process of assigning semantic labels or categories to individual points within a point cloud data structure. A point cloud is a collection of 3D points that represent the surface of an object or a scene, typically acquired through techniques like LiDAR or structured light scanning and the objective of point cloud classification is to extract meaningful information from 3D point cloud data. Point cloud classification has numerous real-world applications, including stereo reconstruction, indoor navigation, autonomous driving, augmented reality, and robotics perception. Compared to 2D image classification, point cloud classification presents additional challenges. While 2D images consist of dense and regular pixel arrays, 3D point clouds are sparse and unordered, lacking the rich texture and information found in 2D images. While early deep learning point cloud classification methods focused on designing advanced architectures for point cloud classification, such as PointNet\cite{qi2017pointnet}, PointNet++ \cite{qi2017pointnet++}, RSCNN \cite{liu2019relation}, DGCNN \cite{wang2019dynamic}, CurveNet \cite{xiang2021walk}, the introduction of Contrastive Language-Image Pre-training (CLIP) has opened up a new direction in computer vision research. CLIP \cite{radford2021learning}, which is trained using unlabeled image-caption supervision, can be adapted to specific downstream tasks with minimal additional training. Inspired by CLIP's \cite{radford2021learning} success, recent works have explored the extension of pre-trained CLIP to 3D recognition. In this direction, CLIP-based point cloud models like PointCLIP \cite{zhang2022pointclip}, CLIP2Point \cite{huang2022clip2point} etc have become state-of-the-art methods in the few-shot setup. In general, the pipeline for 3D recognition using pre-trained CLIP involves projecting the point cloud as depth maps, processing the depth maps using the pre-trained CLIP visual encoder, and adding a small adapter network that is fine-tuned for the specific task.

Although these methods show promising performance for some classes like airplanes, desks, guitars, etc, the performance for some classes like the cup, flower pot, sink, nightstand, etc is still far from satisfactory as depicted by Fig \ref{fig: zero} (b). This is due to the fact that the adapter of CLIP-based models is trained using randomly sampled N-way K-shot data in the standard supervised learning setup. But, for some classes point cloud instances differ drastically in terms of shape and structure as depicted in Fig. \ref{fig: Fig1}. In the CLIP-based point cloud classification few-shot setup, the adapter tries to learn these various shape and structure variations through only k-shot data but often falls short, thus overfitting on the training set and generalizing poorly on evaluation because it is not possible for a single adapter model to perform well on every class-specific point cloud instance. However, another naive approach can be fine-tuning the adapter to the unseen class-specific point cloud instances. But it requires thousands of backpropagation gradients updates, thus requiring considerable time and computation to get the desired result. Moreover, although fine-tuning can solve the unseen class-specific point cloud adaptation problem to some extent, it is not possible to fine-tune the adapter with every unseen class-specific point cloud instance. 

To solve this issue, one potential solution can be the introduction of  a meta-learning formulation for CLIP-based point cloud classification. Unlike supervised learning, meta-learning learns novel tasks with only a few examples, in a similar way to
human beings. Just like humans learn by seeing some examples and then using domain-specific knowledge in practical scenarios, meta-learning adapts to a specific task by observing a few examples. The motivation of meta-learning is to absorb information from one task and generalize that information to unseen tasks proficiently by quickly adapting to the new tasks from a small set of training examples given during the testing phase. Especially, Model Agnostic Meta-learning (MAML) \cite{finn2017model} has shown great impact. MAML \cite{finn2017model} tries to find a good initialization across shared knowledge, so that small updates with task-specific data can adapt the model to that task, thus boosting performance. 

However, the standard MAML \cite{finn2017model} has several problems in this context. Firstly, the standard meta-learning paradigm is used to learn good initialization for the base model which has increased model capacity, enhanced feature extractor, and a large receptive field. In the case of CLIP based model, the adapter is a small model with a smaller receptive field. In the standard MAML \cite{finn2017model}, the base model learns the generalized knowledge from the randomly sampled task, whereas for CLIP-based models, some visual knowledge has already been captured by the CLIP Visual encoder as depicted by the Fig. \ref{fig:zero_few_shot_results} (a). Finally and most importantly, the standard meta-learning gives equal importance to all the classes and samples the N-way K-shot data accordingly. However, in the CLIP-based models, some visual information has already been learned by the CLIP visual encoder (e.g. airplanes, desks, guitars, etc), and the main target of our meta-training is to encode unknown generalized class information (e.g. cup, flower pot, sink, nightstand, etc) to the adapter. Thus, a natural question arises, how can we encode unknown generalized class information to the adapter so that it can achieve satisfactory results?

For this purpose, we propose a novel meta-episodic learning framework with dynamic task sampling for CLIP-based point cloud classification. While meta-episodic learning combines these two concepts by employing episodic training within a meta-learning framework, dynamic task sampling ensures the sampling of unknown classes in the related task within an episodic. For sampling dynamic tasks within an episode, we propose a novel performance memory. The performance memory keeps track of the class-wise performance within an episode and samples the task according. In our approach, the CLIP-based model is trained on episodes of related tasks, where each episode contains a set of related examples mostly sampled from unknown classes. By learning from these episodes, the adapter acquires a generalization ability that enables it to quickly adapt and learn from new tasks. 

In summary, the contribution of this paper is as follows: 1) we, to the best of our knowledge for the first time, propose a novel meta-episodic learning framework for CLIP-based point cloud classification, addressing the challenges of limited training examples and sampling unknown classes. 2) we propose a novel dynamic task sampling technique within the episode based on performance memory. This sampling strategy effectively addresses the challenge of sampling unknown classes, ensuring that the model learns from a diverse range of classes and promotes the exploration of underrepresented categories. By dynamically updating the performance memory, we adaptively prioritize the sampling of classes based on their performance, enhancing the model's ability to handle challenging and real-world scenarios. 3) We confirm that our framework consistently improves upon even the most recent state-of-the-art CLIP-based point cloud models.

\section{Related Works}
\textbf{Deep Learning in Point Clouds: } Deep learning has made significant advancements in point cloud classification and understanding. Following the success of CLIP, several works have emerged that aim to extend the pre-training framework of Contrastive Language-Image Pre-Training (CLIP) to point cloud understanding tasks. For example, Zhang et. al. \cite{zhang2022pointclip} presented PointCLIP to extend CLIP for handling 3D point cloud data along with an interview adapter. Building upon PointCLIP, Zhu et al.  \cite{zhu2022pointclip} introduced PointCLIP V2, which focuses on efficient cross-modal adaptation. They proposed LLM-assisted 3D prompting, which leveraged a language and label mapping (LLM) module to assist in generating informative and compact prompts for point cloud data. Additionally, they introduced realistic shape projection to improve the cross-modal adaptation process. Huang et al. \cite{huang2022clip2point} presented CLIP2Point, which introduced a novel Dual-Path adapter and a contrastive learning framework for transferring CLIP knowledge to the 3D domain.  Yan et al. proposed PointCMT \cite{yan2022let}, a point cloud cross-modal training framework that utilized the advantages of color-aware 2D images and textures to acquire more discriminative point cloud representations. They formulated point cloud analysis as a knowledge distillation problem, transferring knowledge from 2D images to point clouds.

\textbf{Meta Learning: } Meta-learning \cite{finn2017model}, also known as learning to learn, aim to acquire prior knowledge or meta-knowledge from the distribution of tasks and use this knowledge to facilitate learning on new, unseen tasks. In this direction, MAML \cite{finn2017model} aims to learn a good initialization of model parameters that can be easily adapted to new tasks with a few gradient steps. The key idea behind MAML \cite{finn2017model} is to optimize the model's parameters in a way that they can be fine-tuned quickly to minimize the loss of new tasks. MAML++ \cite{antoniou2018train} is an extension of the original MAML \cite{finn2017model} algorithm that introduces several improvements to enhance its performance. One key enhancement is the use of a second-order approximation for computing gradients, which takes into account the curvature of the loss landscape. By considering second-order information, MAML++ \cite{antoniou2018train} can capture more nuanced relationships between model parameters and achieve better adaptation to new tasks. MetaSGD \cite{li2017meta} is another optimization-based meta-learning algorithm that addresses the challenges of learning an effective optimization algorithm itself. The main idea behind MetaSGD \cite{li2017meta} is to learn a meta-optimizer that dynamically adjusts the learning rate or optimization strategy for different tasks. Instead of learning model parameters directly, MetaSGD \cite{li2017meta} aims to learn the update rule or optimization algorithm that adapts the model's parameters.

\begin{figure*}[t]
\centering
\includegraphics[width=0.85\linewidth]{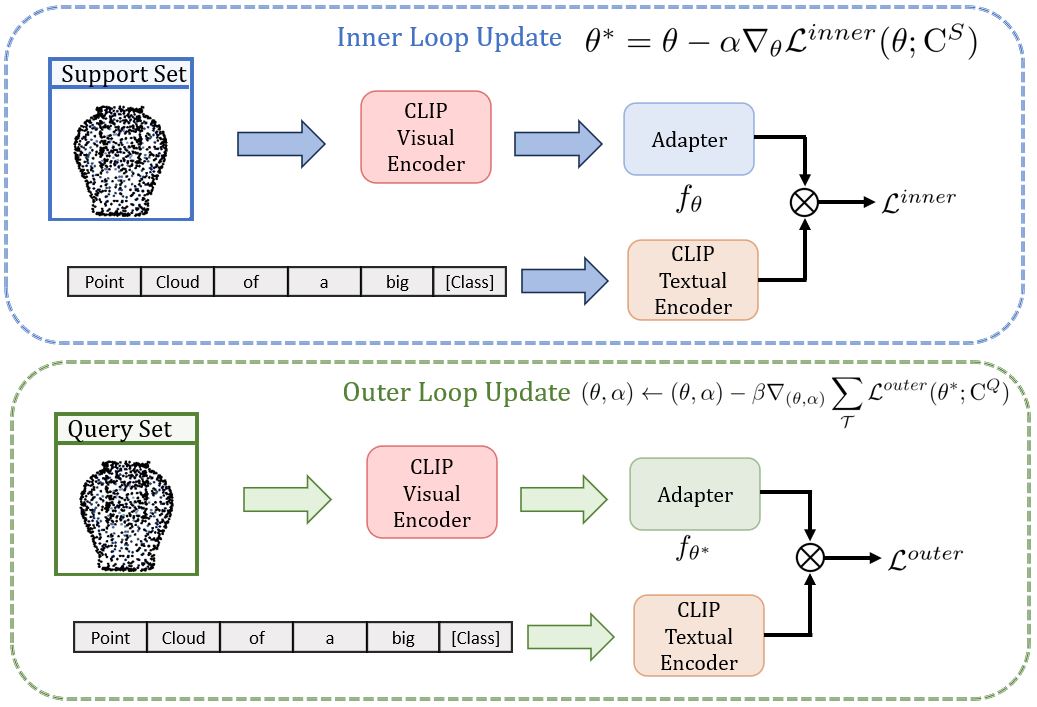}
\vspace{-0.20cm}
\caption{The adapter learns following a bi-level optimization process. While the adapter learns to recognize and discriminate the features of the inner loop update, the outer loop extracts meta-features of the point clouds that generalize across tasks. Additionally, we introduce dynamic task sampling within the episode based on performance memory to ensure underrepresented class sampling.}
\label{fig:adapter_process}
\vspace{-0.50cm}
\end{figure*}

\textbf{Sampling in Fewshot Learning: }  Sampling plays a crucial role in few-shot learning by shaping the training and evaluation processes, enabling effective adaptation and generalization from limited labeled examples. Several works have been proposed in the last few years in this direction. For example, Pezeshkpour et. al. \cite{pezeshkpour2020utility}  investigated the effectiveness of active instance selection in the context of few-shot learning and  proposed a framework where instances were selected based on their informativeness, relevance, and diversity. Arnold et.al. \cite{arnold2021uniform}  highlighted the importance of considering episode difficulty and proposed a uniform sampling strategy to address this factor in few-shot learning. Le et. al. \cite{le2021poodle} introduced a method called Poodle that aimed to enhance few-shot learning performance by addressing the challenge of out-of-distribution samples. Xu et. al. \cite{xu2022generating} proposed a method for generating representative samples to improve the few-shot classification. Xu et. al. \cite{xu2022alleviating} addressed the issue of sample selection bias in few-shot learning and proposed a method to alleviate this bias by removing the projection to the centroid. Roy et. al. \cite{roy2022felmi} introduced a method called FeLMi that improved few-shot learning by incorporating a technique called hard mixup. The hard mixup technique involved selecting difficult samples from the support set and blending them with other samples from different classes.

\section{Methodology}
In this section, we first briefly describe baseline CLIP-based point cloud models like PointCLIP \cite{zhang2022pointclip}, CLIP2Point \cite{huang2022clip2point} for point cloud classification (Sec. \ref{baseline}). Then we introduce our meta-episodic learning framework for point cloud classification (Sec.\ref{meta}). Finally, we describe our proposed dynamic task sampling technique based on performance memory for class instance adaptive point cloud classification (Sec. \ref{dynamic}). The overall overview of our method is depicted in Fig.~\ref{fig:adapter_process}.

\begin{figure*}[t]
\centering
\includegraphics[width=0.85\linewidth]{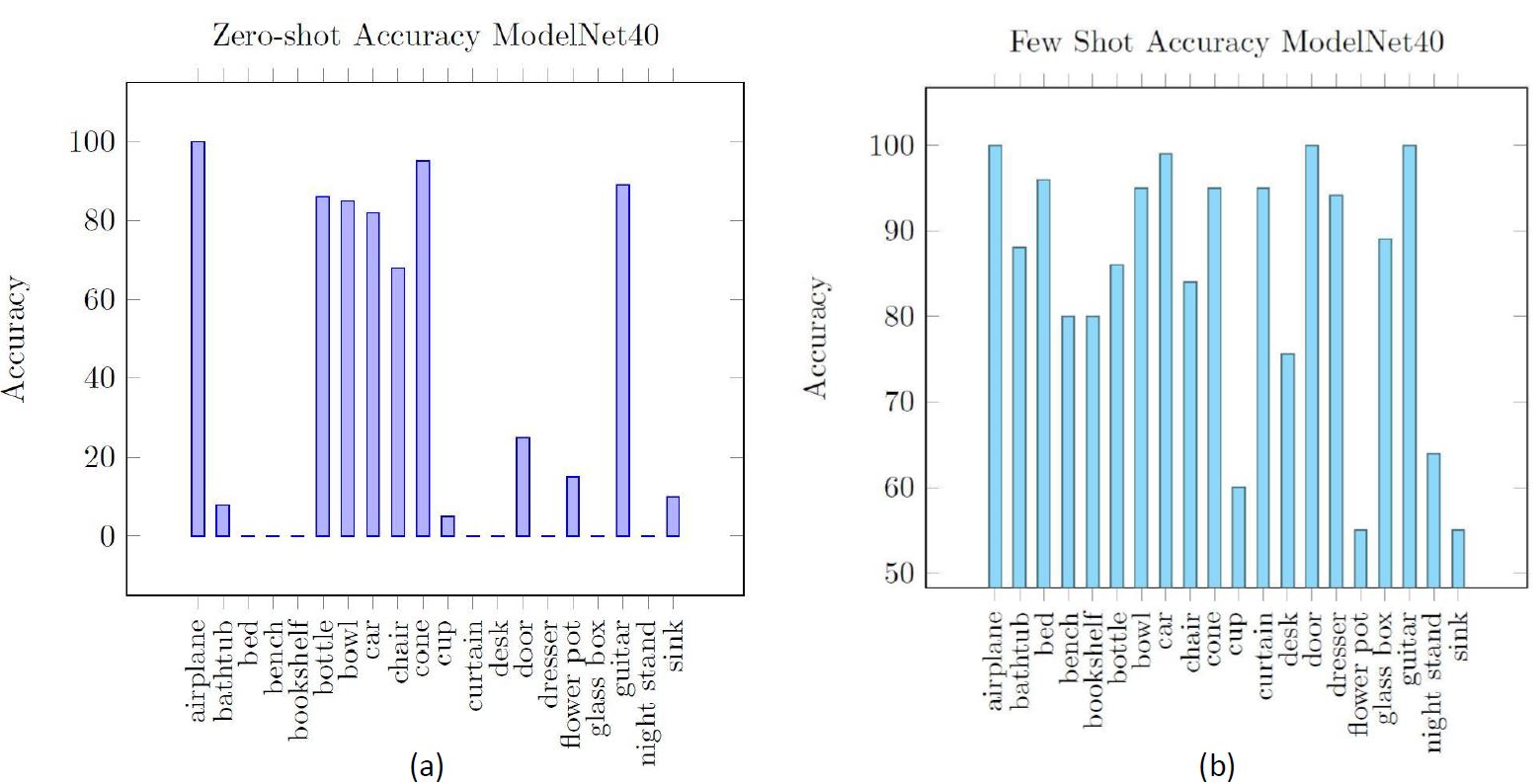}
\vspace{-0.20cm}
\caption{(a) Zero-shot result of PointCLIP \cite{zhang2022pointclip}. We can observe that CLIP's Visual encoder has already captured certain classes such as airplanes, desks, guitars, etc. (b) Few-shot result of PointCLIP \cite{zhang2022pointclip}. Some classes like the cup, flower pot, sink, nightstand, etc., exhibit lower performance in the few-shot setup.}
\label{fig:zero_few_shot_results}
\vspace{-0.60cm}
\end{figure*}

\subsection{Baseline CLIP-based point cloud models}\label{baseline}
The CLIP-based point cloud models are the state-of-the-art point cloud classification methods in the few-shot setup. In this paper, we select two CLIP-based point cloud methods, PointCLIP \cite{zhu2022pointclip} and CLIP2Point \cite{huang2022clip2point}, to use as our baseline CLIP-based point cloud models. 

For completeness, we briefly summarize the outline of the CLIP-based point cloud models. In general, they consist of three components: (a) a pre-trained CLIP visual Encoder, $F_V$ (b)  a pre-trained CLIP textual Encoder $F_T$ (c) a small trainable network called an adapter $F_{\theta}$. For $N$ class classification, CLIP-based models use a pre-defined template: ``point cloud depth map of a [CLASS]" and the textual encoder outputs $W_t \in R^{N \times C}$, where $C$ is the channel of the text embedding. To feed the Point cloud to the CLIP's visual encoder, point clouds are first projected into $\{f_1,f_2, \dots, f_m\}$ depth maps where $m$ denotes the number of views and $f_i \in W^{1 \times C}$ denotes each view of the point cloud. Given the input $\{f_1,f_2, \dots, f_m\}$, the CLIP's visual encoder generates visual feature $\{F_1^I,F_2^I, \dots, F_m^I\}$, where $F_i^I \in R^{1 \times C}$, $C$ being the channel information of the embedding. In the zero-shot setup, there is no training stage and no adapter is required. Each viewpoint generates a prediction by calculating the cosine similarity between the visual feature $F_i^I$ and the  textual feature $F^T$. The final prediction is a weighted sum of all viewpoint-wise predictions. 

For few-shot point cloud classification, the adapter $F_{\theta}$ is trained using randomly sampled N-way K-shot data from the training dataset in the standard supervised learning setup. For this purpose, PointCLIP \cite{zhang2022pointclip} proposes an interview adapter that extracts a global visual representation by combining multi-view features generated by the CLIP visual encoder. The global representation obtained from the inter-view adapter serves as a higher-level understanding of the visual input and encapsulates important information that is shared across different views. During inference, CLIP's visual feature $\{F_1^I,F_2^I, \dots, F_m^I\}$ is passed through the trained adapter network $F_{\theta}$ and the final prediction is calculated as cosine similarity between the output adapter feature and the textual feature. 

On the other hand, CLIP2Point \cite{huang2022clip2point} proposes a Dual-Path Adapter (DPA) that combines the benefits of pre-training with instance-level depth maps and category-level discrimination from CLIP pre-training knowledge. The DPA module consists of a dual-path structure, utilizing two encoders: the pre-trained depth encoder and the CLIP visual encoder. The combination of the two encoders allows for a comprehensive representation of the visual data, incorporating both instance-level depth information and category-level discrimination from CLIP pre-training. By combining the outputs of the two encoders, the DPA module generates the final logits. We refer the reader to \cite{zhu2022pointclip, huang2022clip2point} for further architectural details. In this paper, the adapter indicates the interview adapter for PointCLIP and the dual-path adapter for CLIP2Point and denotes the complete set of parameters as $\theta$.

\subsection{Meta Episodic Learning for Point Cloud Classification} \label{meta}
Traditionally, the CLIP-based point cloud model inputs a point cloud $\rho$ and generates its corresponding class label $Y$. Conventional CLIP-based point cloud models learn from $N$ way $K$ shot data, where N denotes the total number of classes in the dataset and K notes the total number of instances per class. Due to the data instance-specific training, it ignores the class-specific shape and structure variation distribution. Henceforth, the performance deteriorates on some point cloud classes because of poor generalization on diverse point cloud class instances.

In contrast, we take a meta-episodic learning approach for the few-shot learning of the adapter along with dynamic tasks to address the challenge of CLIP-based point cloud classification. It combines the concepts of meta-learning and episodic training to enable efficient adaptation to new tasks with limited training examples. The goal is to encode unknown generalized class information into the adapter, allowing it to achieve satisfactory results. Let's consider $W_{tr}$ and $W_{test}$ denote the disjoint training and testing point cloud set respectively, i.e., $W_{tr} \cap W_{test} = \phi $. The training and testing set are denoted as $\mathcal{C}^{tr}=\{\mathcal{C}^{tr}_1, \mathcal{C}^{tr}_2, \dots, \mathcal{C}^{tr}_{|W_{tr}|}\}$ and $\mathcal{C}^{test}=\{\mathcal{C}^{test}_1, \mathcal{C}^{test}_2, \dots, \mathcal{C}^{test}_{|W_{test}|}\}$. Each $i$-th point cloud class in the training set has  its own total of $M$ labeled point cloud instances as $\mathcal{C}_{i} = \{ (\rho_1, Y), (\rho_2, Y), \cdots, (\rho_{M}, Y) \}$. In the meta-episodic training, let each episode consists of $S$ tasks, where each $\mathcal{T}$ consists of a support set $\mathrm{C}^{S}$ and a query set $\mathrm{C}^{Q}$. Intuitively, our approach tries to find a good initialization of parameters $\theta$, representing the prior or generalized knowledge of point clouds, so that a few updates of $\theta$ using $\mathrm{C}^{S}$ can make large improvements by reducing the error measures and boosting the performance in $\mathrm{C}^{Q}$. To learn this optimal initialisation parameter $\theta$, we first adapt (task-specific) $f_{\theta}$ using $\mathrm{C}^{S}$ by fine-tuning:
\begin{equation}
    \theta^{*} = \theta - \alpha \nabla_\theta \mathcal{L}^{inner}(\theta; \mathrm{C}^{S})
\end{equation}

Where $\mathcal{L}^{inner}$ denotes loss in the inner loop and $\alpha$ denotes inner loop learning rate. In the inner loop update, the adapter learns to recognize and discriminate the patterns, features, and characteristics relevant to the point cloud classification task. Evaluation of the adapted model is performed on unseen examples sampled from the same task $\mathrm{C}^{Q}$, to measure the generalization of $f_{\theta^{*}}$. This acts as feedback for MAML \cite{finn2017model} to adjust its initialization parameters $\theta$ to achieve better generalization on any $\mathcal{T}$ (across-task):
\begin{equation}
     (\theta, \alpha) \leftarrow (\theta, \alpha) - \beta \nabla_{(\theta, \alpha)} \sum_{\mathcal{T}} \mathcal{L}^{outer}(\theta^{*} ; \mathrm{C}^{Q})
\end{equation}
Where $\mathcal{L}^{outer}$ denotes loss in the outer loop and $\beta$ denotes outer loop learning rate. The primary objective of the outer loop is to extract meta-knowledge or meta-features that generalize across tasks. However, the standard meta-learning has several limitations for the CLIP-based Point Cloud classification. To address these challenges, we propose novel dynamic sampling within the episode based on performance memory.

\subsection{Dynamic Task Sampling} \label{dynamic}
The standard Model Agnostic Meta-learning (MAML) \cite{finn2017model} is not well-suited for CLIP-based models due to several reasons. Firstly, the standard meta-learning paradigm is used to learn good initialization for the base model which has increased model capacity, enhanced feature extractor, and a large receptive field. In the case of CLIP based model, the adapter is a small model with a smaller receptive field. In the standard MAML \cite{finn2017model}, the base model learns the generalized knowledge from the randomly sampled task, whereas for CLIP-based models, some visual knowledge has already been captured by the CLIP Visual encoder as depicted by the Fig. \ref{fig:zero_few_shot_results} (a). Finally and most importantly, the standard meta-learning gives equal importance to all the classes and samples the N-way K-shot data accordingly. However, in the CLIP-based models, some visual information has already been learned by the CLIP visual encoder (e.g. airplanes, desks, guitars, etc), and the main target of our meta-training is to encode unknown generalized class information (e.g. cup, flower pot, sink, nightstand, etc) to the adapter.

\setlength{\textfloatsep}{4pt}
\begin{algorithm}[!hbt]
\caption{Training of our approach}
\label{algo:train}
\sloppy
\begin{algorithmic}[1]
    \State \textbf{Input}: Training dataset $\mathcal{C}^{tr}=\{\mathcal{C}^{tr}_1, \mathcal{C}^{tr}_2, \dots, \mathcal{C}^{tr}_{|W_{tr}|}\}$; $\beta$ as learning rate.
    \State \textbf{Requires}: Pre-trained CLIP visual encoder $F_V$, Pre-trained CLIP textual encoder $F_T$
    \State \textbf{Initialise}: Initialise Adapter $\theta$, inner learning rate $\alpha$
    \State \textbf{Output}: Optimised meta-parameters $\{\theta, \alpha\}$
    \ForEach {episode}
        \State Initialize Performance Memory $P$ with keys as the class labels and values as 0
        \For{task = $1,2, \dots, S$}
            \State Sort $P$ based on values and decide classes to sample
            \State Sample task $\mathcal{T} = \{\mathrm{C}^{S}, \mathrm{C}^{Q}\}$ from the decided classes
            \ForEach {task $\mathcal{T}$}
                \State Evaluate inner objective: $\mathcal{L}^{inner}(\theta; \mathrm{C}^{S})$
                \State Adapt: $\theta^{*} = \theta - \alpha \nabla_\theta \mathcal{L}^{inner}(\theta; \mathrm{C}^{S})$
                \State Compute outer objective: $\mathcal{L}^{outer}(\theta^{*}; \mathrm{C}^{Q})$
            \EndFor
            \State Update meta-parameters: $(\theta, \alpha) \leftarrow (\theta, \alpha) - \beta \nabla_{(\theta, \alpha)} \sum_{\mathcal{T}} \mathcal{L}^{outer} (\theta^{*} ; \mathrm{C}^{Q})$
            \State Update the class-wise accuracy in $P$
        \EndFor
    \EndFor
\end{algorithmic}
\end{algorithm}

To address these challenges, we propose novel dynamic sampling within the episode based on performance memory.  Dynamic task sampling is a crucial component of the proposed meta-episodic learning framework for CLIP-based point cloud classification and performance memory plays a critical role in guiding the task sampling process. It aims to address the challenge of effectively sampling unknown classes within episodes, allowing the model to learn from diverse and relevant data. Let $P$ denote performance memory and $P[i]$  denote $i-th$ class in the performance memory. At the onset of each episode, we initialize all the $P[i]$ equal to 0. For task sampling within an episode, we employ a specific procedure. Initially, we sort the performance memory in ascending order based on the recorded values. This sorting enables us to prioritize the classes based on their performance. Consequently, we select the first N classes from the performance memory, facilitating an N-way task sampling. Throughout the episode, the performance memory is updated dynamically as follows:

\begin{equation}
     P[i] = \frac{P[i] + A_i}{2}
\end{equation}

Where $A_i$ denotes $i-th$ class accuracy in the query set $\mathrm{C}^{Q}$ of a task. This updating process allows the performance memory to reflect the evolving performance of each class within the episode. By incorporating dynamic sampling within the episode based on performance memory, our proposed approach ensures that the model is exposed to a diverse range of classes during training. It prioritizes the sampling of classes that have performed relatively poorly, promoting the model's ability to learn from challenging and underrepresented classes.

\begin{algorithm}[!hbt]
\sloppy
	\caption{Inference of our model}
	\label{algo:infer}
	\begin{algorithmic}[1] 
        \State \textbf{Input}: Testing dataset $\mathcal{C}^{test}=\{\mathcal{C}^{test}_1, \mathcal{C}^{test}_2, \dots, \mathcal{C}^{test}_{|W_{test}|}\}$; number of gradient updates $n$. 
        \State \textbf{Requires}: Pre-trained CLIP visual encoder $F_V$, Pre-trained CLIP textual encoder $F_T$, Optimised meta-parameters $\{\theta, \alpha\}$
            \State Sample task $\mathcal{T} = \{\mathrm{C}^{S}, \mathrm{C}^{Q}\}$ from $\mathcal{C}^{test}$
            \For {$n$ steps}
                \State Evaluate inner objective: $\mathcal{L}^{inner}(\theta; \mathrm{C}^{S})$
                \State Adapt: $\theta_{c} = \theta - \alpha \nabla_\theta \mathcal{L}^{inner}(\theta; \mathrm{C}^{S})$
            \EndFor
            \State \textbf{Return} class instance specialised Adapter params. $\mathbf{\theta_{c}}$.
	\end{algorithmic} 
\end{algorithm}

In summary, our meta-episodic learning framework leverages dynamic sampling within the episode, guided by the performance memory. This approach effectively addresses the challenge of sampling unknown classes, enabling the model to learn from diverse and relevant data. The updating of the performance memory ensures that the model adapts to the evolving performance of different classes, enhancing its ability to handle real-world scenarios and achieve improved performance in CLIP-based point cloud classification tasks. The training and inference process of our approach is summarised in Algorithm \ref{algo:train} and \ref{algo:infer}, respectively.

\section{Experiments}

\textbf{Datasets:} We evaluate our proposed approach on two widely used benchmark datasets ModelNet40 \cite{wu20153d} and ScanObjectNN \cite{uy2019revisiting}. ModelNet40 is a synthetic indoor 3D dataset that consists of 40 classes of point cloud objects. It has a training set of 9,843 and a testing set of 2,468 point clouds respectively. As the original ModelNet40 is not aligned in orientation, in our approach we use the aligned version of ModelNet 40 \cite{sedaghat2016orientation}. ScanObjectNN is a real-world point cloud dataset that has a training set of 2,321 and a testing set of 581 samples from 15 point cloud classes. Following PointCLIP \cite{zhang2022pointclip} and CLIP2Point \cite{huang2022clip2point}, we train our approach on the ModelNet40 training set and test on ModelNet40 and  ScanObjectNN testing set.

\textbf{Implementation Details:} We implemented our framework in PyTorch \cite{paszke2019pytorch} and conducted experiments on a 12 GB Nvidia Titan X GPU. We train the adapter from scratch with 100 epochs, and each epoch contains 50 meta-training
episodes according to algorithm \ref{algo:train}. Each episode consists of 20 tasks and each task consists of N classes with K-labeled support examples and Q query examples for each class, which is denoted as the N-way K-shot Q-query setting. In our approach, we use a 3-way 5-shot 5-query for our tasks. Once the meta-training is ended, we test the network according to algorithm \ref{algo:infer}. For meta-training and meta-testing, we used a one-step gradient update. We use ADAM as a meta-optimizer with outer-loop learning rate $\beta$ as 0.0001, while the inner-loop learning $\alpha$ is meta-learned during training. We use accuracy as our evaluation metric.

\subsection{Competitors}

To the best of our knowledge, no prior research has specifically addressed task-specific few-shot learning for CLIP-based Point cloud models. However, in order to validate our approach, we have designed four strong baselines from the perspective of optimization-based meta-learning. These baselines are as follows: \textbf{(i) MAML \cite{finn2017model}: } In this baseline, we train CLIP-based models using standard random task sampling, without employing episodes. \textbf{(ii) Reptile \cite{nichol2018reptile}:  } This baseline involves directly performing inner loop updates using a randomly sampled support set of tasks. The models converge to initialization by accumulating changes made during these updates across multiple tasks. \textbf{(iii) MetaSGD \cite{li2017meta}: } Here, CLIP-based models are trained within a meta-episodic learning framework along with random task sampling, where each parameter has its own learning rate.  \textbf{(iv) Ours: } We train CLIP-based models using a MAML approach within a meta-episodic learning framework along with dynamic task sampling.

\begin{table}[t]
\vspace{-0.10cm}
\centering
\scriptsize
\caption{Comparison (\%) of baseline and our approach on PointCLIP and CLIP2Point using the prompt "point cloud of a big [CLASS]".}
\vspace{-0.20cm}
\begin{tabular}{c|c|c|c|c}
\hline
\textbf{Methods} & \multicolumn{2}{c|}{\textbf{ModelNet40}} & \multicolumn{2}{c}{\textbf{ScanobjectNN}} \\
\hline
 & \textbf{Baseline} & \textbf{Ours} & \textbf{Baseline} & \textbf{Ours} \\
\hline
PointCLIP & 83.80 & \textbf{86.93} & 54.37 & \textbf{58.72} \\
\hline
CLIP2Point & 85.10 & \textbf{88.64} & 57.49 & \textbf{63.65} \\
\hline
\end{tabular}

\label{tab:comparison_results}
\end{table}

\subsection{Result Analysis and Discussion}
Table \ref{tab:comparison_results} compares the baseline results with our proposed approach on PointCLIP \cite{zhang2022pointclip} and CLIP2Point \cite{huang2022clip2point} models using the prompt "point cloud of a big [CLASS]."  For ModelNet40, our approach has outperformed the baseline  model by 3.13\% and 3.54\% on PointCLIP \cite{zhang2022pointclip} and  CLIP2Point \cite{huang2022clip2point} respectively. Similarly, on the ScanobjectNN dataset, the performance improves by 4.35\% and 6.16\% 
on PointCLIP \cite{zhang2022pointclip} and CLIP2Point \cite{huang2022clip2point} respectively. This result indicates the robustness of our meta-episodic approach compared to the standard few-shot approach used by CLIP-based point cloud models. 
\begin{table}[t]
\vspace{-0.10cm}
\centering
\scriptsize
\caption{Performance (\%) analysis with different Baselines on ModelNet40 and ScanobjectNN using prompt “point cloud of a big [CLASS]”.}
\vspace{-0.20cm}
\begin{tabular}{c|c|c}
\hline
\textbf{Methods} & \textbf{ModelNet40} & \textbf{ScanobjectNN} \\
\hline
PointCLIP Baseline & 83.80 & 54.37 \\
PointCLIP + MAML & 84.78 & 55.69 \\
PointCLIP + Reptile & 84.10 & 55.04 \\
PointCLIP + MetaSGD & 84.92 & 55.85 \\
PointCLIP + Ours & \textbf{86.93} & \textbf{58.72} \\
\hline
CLIP2Point Baseline & 85.10 & 57.49 \\
CLIP2Point + MAML & 86.34 & 58.56 \\
CLIP2Point + Reptile & 85.83 & 58.24 \\
CLIP2Point + MetaSGD & 86.36 & 58.68 \\
CLIP2Point + Ours & \textbf{88.64} & \textbf{63.65} \\
\hline
\end{tabular}
\label{tab:Performance}
\end{table}

Table \ref{tab:Performance} presents a performance analysis with different Competitors on ModelNet40 and ScanobjectNN datasets. The Competitors include PointCLIP \cite{zhang2022pointclip} and CLIP2Point \cite{huang2022clip2point} with various meta-learning algorithms such as MAML, Reptile, and MetaSGD, as well as our proposed approach. For PointCLIP \cite{zhang2022pointclip}, our proposed approach outperforms MAML by 2.15\% and 3.03\% on the ModelNet40 and ScanobjectNN datasets, respectively. Similarly, our approach surpasses Reptile by 2.83\% and 3.68\%, and MetaSGD by 2.01\% and 2.87\% on the same datasets. For CLIP2Point \cite{huang2022clip2point}, our approach demonstrates superior performance compared to the baselines. It outperforms MAML by 2.3\% and 5.09\% on the ModelNet40 and ScanobjectNN datasets, respectively. Moreover, our model surpasses Reptile by 2.81\% and 5.41\%, and MetaSGD by 2.28\% and 4.97\% on the respective datasets. These results highlight the effectiveness of our proposed approach in achieving better performance compared to the other competitors in the task-specific few-shot learning setting for CLIP-based point cloud models. 

\begin{table}[t]
\vspace{-0.10cm}
\centering
\small 
\setlength{\tabcolsep}{1pt} 
\caption{Comparison (\%) of  Baseline and our approach on PointCLIP for different prompt designs on ModelNet40. }
\vspace{-0.20cm}
\begin{tabular}{c |c| c} 
 \hline
Prompts & PointCLIP  &  PointCLIP  \\ [0.5ex] 
 & Baseline  &  + Ours \\
 \hline
“a photo of a [CLASS].” & 81.78 & \textbf{84.55} \\ 
 \hline
“a point cloud photo of a [CLASS].” & 82.02 & \textbf{86.12}  \\ 
 \hline
“point cloud of a [CLASS].” & 82.10  &  \textbf{86.82} \\ 
 \hline
“point cloud of a big [CLASS].” & 83.80  & \textbf{86.93} \\ 
 \hline
“point cloud depth map of a [CLASS].” & 81.58  & \textbf{85.97} \\ 
 \hline

\end{tabular}
\label{tab:prompt_comparison}
\end{table}

Table \ref{tab:prompt_comparison} focuses on different prompt designs for the PointCLIP \cite{zhang2022pointclip} model on the ModelNet40 dataset. It compares the baseline accuracy with our approach's accuracy for various prompt designs. Our approach consistently outperforms the baseline in all prompt designs.

\subsection{Ablation Study}

\textbf{The effect of Dynamic task sampling: }  To observe the effect of dynamic task sampling, we conduct an experiment with dynamic task sampling and random task sampling  using the prompt “point cloud of a big [CLASS]” on ModelNet40 as shown in Tab.~\ref{tab:dynamic}. 

\begin{table}[t]
\vspace{-0.10cm}
\centering
\caption{Effect of dynamic task sampling on ModelNet40 using prompt “point cloud of a big [CLASS]”. }
\vspace{-0.20cm}
\begin{tabular}{ccc}
\hline
 & PointCLIP & CLIP2Point  \\ \hline
Random task sampling & 84.78 & 86.34 \\
Dynamic task sampling & \textbf{86.93} &\textbf{88.64}\\ \hline
\end{tabular}
\label{tab:dynamic}
\end{table}

From the table, it is evident that incorporating  dynamic task sampling on the few-shot pipeline improves accuracy by 2.15 \%. and 2.3 \% for PointCLIP \cite{zhu2022pointclip} and CLIP2Point \cite{huang2022clip2point} respectively. This is because dynamic task sampling prioritizes the sampling of classes that have performed relatively poorly, promoting the model’s ability to learn from challenging and underrepresented classes.

\begin{figure}[hbt]
\vspace{-0.10cm}
\begin{center}
\includegraphics[width=\linewidth]{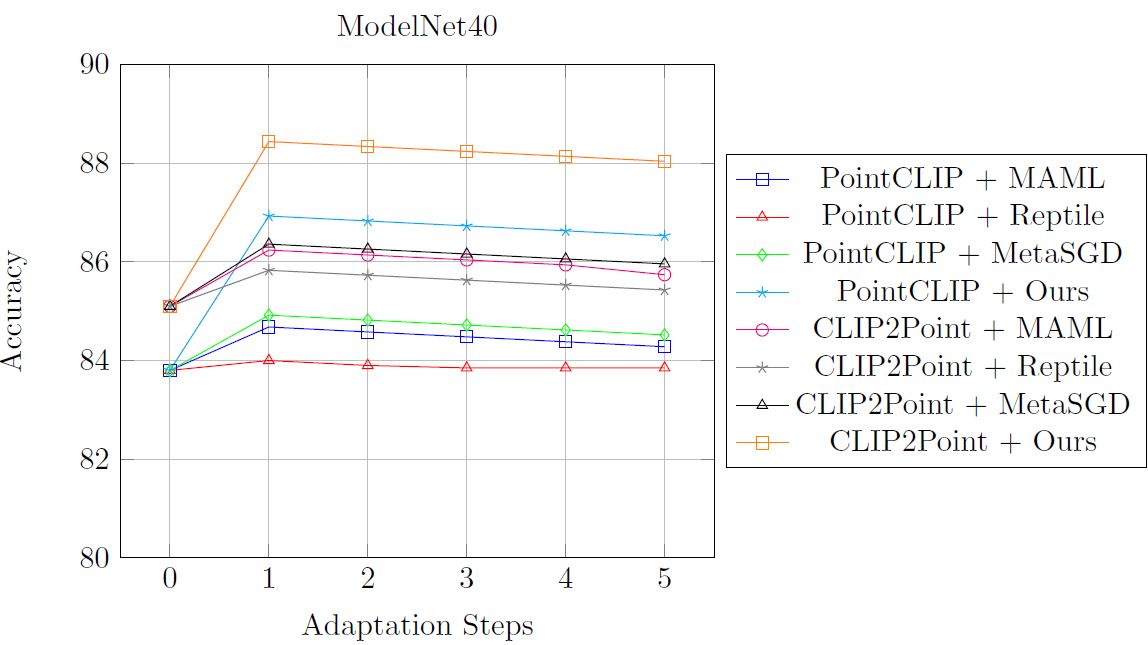}
\end{center}
\vspace{-.40cm}
  \caption{Accuracy of competitors with varying adaptation steps on ModelNet40 using prompt “point cloud of a big [CLASS]”.}
\label{fig:fig4}
\end{figure}

\textbf{Number of Adaptation Steps: } In Fig. \ref{fig:fig4}, we experiment with varying the number of adaptation steps during the inference process. we observe that using just a single gradient step update, which is the most common approach in our experiments, yields the highest performance gain. The reason behind this result could be that the inner loop, which handles the adaptation process, focuses too much on capturing unnecessary point cloud style details, causing it to forget the generic prior knowledge learned by the model. As a result, performing additional updates in the inner loop might lead to diminishing results, as the model becomes overfit to the specific style of the training examples and loses its ability to generalize well to new tasks.

\section{Conclusion}
In conclusion, we have proposed a novel meta-episodic learning framework for CLIP-based point cloud classification, addressing the challenges of limited training examples and sampling unknown classes. Our framework combines meta-learning and episodic training, enabling the model to quickly adapt and generalize to new tasks. Additionally, we have introduced dynamic task sampling within the episode based on performance memory. This sampling strategy effectively addresses the challenge of sampling unknown classes, ensuring that the model learns from a diverse range of classes and promotes the exploration of underrepresented categories. By dynamically updating the performance memory, we adaptively prioritize the sampling of classes based on their performance, enhancing the model's ability to handle challenging and real-world scenarios.



\bibliographystyle{IEEEtranBST/IEEEtran}
\bibliography{IEEEtranBST/IEEEfull}
%



\end{document}